\documentclass[letterpaper, 10pt, conference]{ieeeconf}

\IEEEoverridecommandlockouts
\usepackage{graphics}
\usepackage{epsfig}
\usepackage{times}
\usepackage{amsmath}
\usepackage{amssymb}

\usepackage[bottom]{footmisc}

\usepackage{mathtools}
\DeclarePairedDelimiter\abs{\lvert}{\rvert}

\usepackage{textcomp}

\makeatletter
\let\MYcaption\@makecaption
\makeatother

\usepackage[font=footnotesize,aboveskip=12pt]{subcaption}
\makeatletter
\let\@makecaption\MYcaption
\makeatother

\makeatletter
\let\NAT@parse\undefined
\makeatother

\usepackage[noadjust]{cite}
\usepackage{hyperref}
\usepackage{cleveref}
\Crefname{equation}{Eq.}{Eqs.}
\Crefname{figure}{Fig.}{Figs.}
\Crefname{tabular}{Tab.}{Tabs.}
\crefname{appsec}{Appendix}{Appendices}

\newcommand{\isep}{\mathrel{{.}\,{.}}\nobreak}

\title{\LARGE\bf Neuromorphic control for optic-flow-based landings of MAVs \\using the Loihi processor}

\author{Julien Dupeyroux, Jesse J. Hagenaars, Federico Paredes-Vall\'{e}s and Guido C.\thinspace H.\thinspace E. de Croon
    \thanks{This project has received funding from the ECSEL Joint Undertaking (JU) under grant agreement No. 826610. The JU receives support from the European Union's Horizon 2020 research and innovation program and Spain, Austria, Belgium, Czech Republic, France, Italy, Latvia, Netherlands.}
    \thanks{All authors are with the Micro Air Vehicle Lab, Faculty of Aerospace Engineering, Delft University of Technology, The Netherlands. Contact: \href{mailto:j.j.g.dupeyroux@tudelft.nl}{\texttt{j.j.g.dupeyroux@tudelft.nl}}}
}

\begin{document}

\maketitle
\thispagestyle{empty}
\pagestyle{empty}

\begin{abstract}
Neuromorphic processors like Loihi offer a promising alternative to conventional computing modules for endowing constrained systems like micro air vehicles (MAVs) with robust, efficient and autonomous skills such as take-off and landing, obstacle avoidance, and pursuit. However, a major challenge for using such processors on robotic platforms is the reality gap between simulation and the real world. In this study, we present for the very first time a fully embedded application of the Loihi neuromorphic chip prototype in a flying robot. A spiking neural network (SNN) was evolved to compute the thrust command based on the divergence of the ventral optic flow field to perform autonomous landing. Evolution was performed in a Python-based simulator using the PySNN library. The resulting network architecture consists of only 35 neurons distributed among 3 layers. Quantitative analysis between simulation and Loihi reveals a root-mean-square error of the thrust setpoint as low as 0.005~g, along with a 99.8\% matching of the spike sequences in the hidden layer, and 99.7\% in the output layer. The proposed approach successfully bridges the reality gap, offering important insights for future neuromorphic applications in robotics. Supplementary material is available at \url{https://mavlab.tudelft.nl/loihi/}.
\end{abstract}
\vspace{0.1cm}

\section{Introduction}

\begin{figure}
    \centering
    \includegraphics[width=\linewidth]{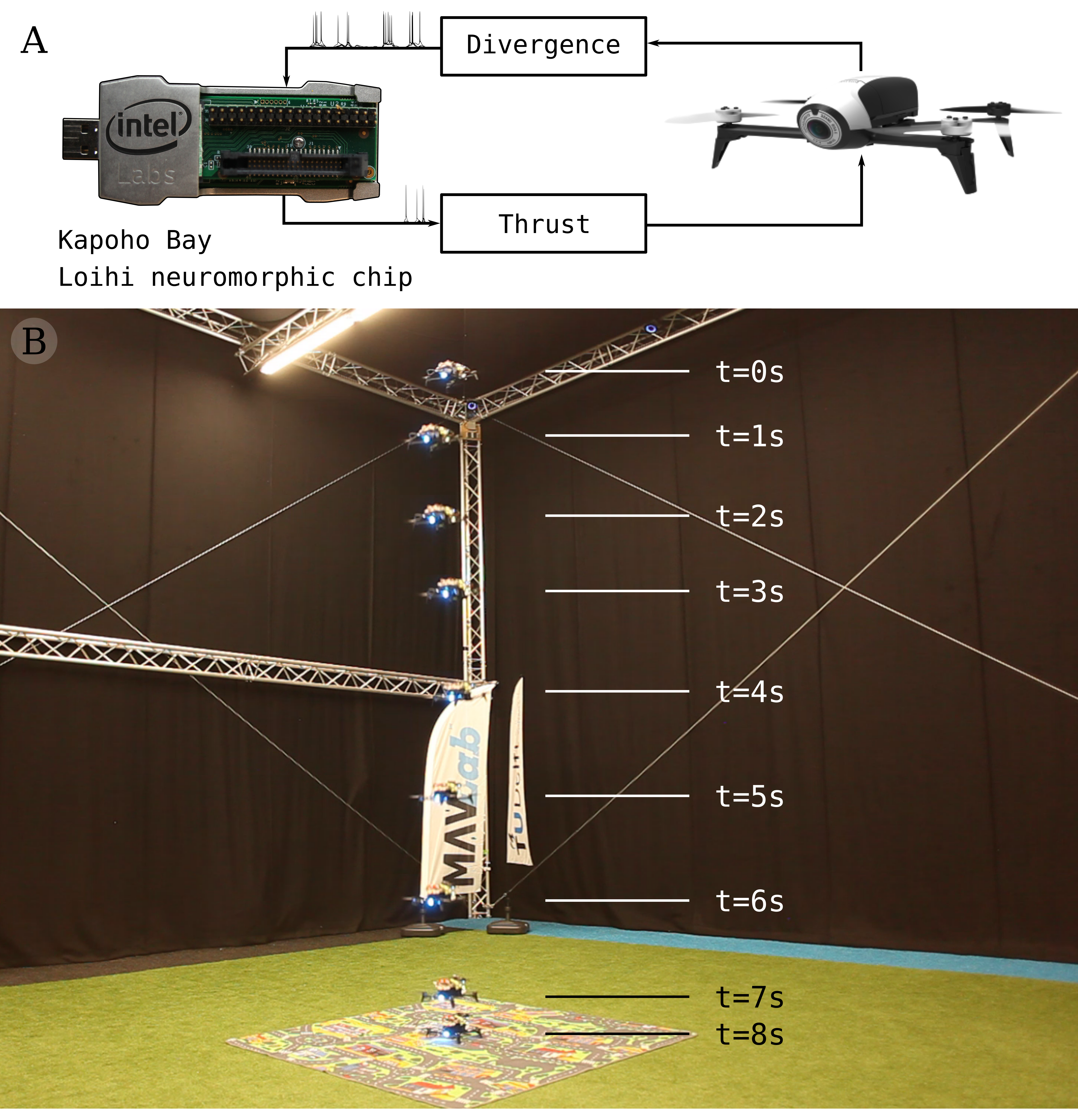}
    \caption{Optic flow landing experiment with the Loihi neuromorphic chip in the control loop. \textbf{A:} Architecture of the neuromorphic control of the MAV's thrust to achieve autonomous landing. The Kapoho Bay USB stick features two Loihi neuromorphic chips for a total of approximately 262,000 neurons and 260 million synapses. \textbf{B:} Visualization of a landing test in our flying arena. The thrust setpoint is adjusted by the SNN running on the Loihi chip in response to the divergence error of the optic flow field.}
    \label{fig:main}
\end{figure}

Providing micro air vehicles (MAVs) with complete autonomy is a complex challenge that generally requires multiple sensors and sensory redundancy along with significant computational resources, which are constraints that do not always fit with MAVs. Yet, flying insects like fruit flies, with their $\sim$100,000 neurons, are known to excel in flying in unknown and complex environments, performing fast maneuvers, avoiding obstacles, chasing mates, and navigating~\cite{zheng2018complete}. To achieve such a performance, these insects rely on visual cues like the optic flow, i.e. the brightness change in the retina caused by the relative motion of the observer~\cite{srinivasan2004visual, serres2017optic}. It was recently demonstrated that hoverflies use visual cues to stabilize their flight during free-fall to avoid a crash~\cite{goulard2016crash}. Besides, honeybees maintain a constant divergence (rate of expansion) of the optic flow to achieve smooth landing~\cite{baird2013universal}. Given the very low number of neurons in the flying insect brain, we must acknowledge that they represent a gold standard in autonomous flight, one that we should take inspiration from to design robust and efficient autonomous flight controllers.

Like any other animal, insects process visual information in a spike-based manner by means of a densely connected neural structure that goes from the compound eye to the optic lobe, which is involved in the optic flow estimation, and then to the central complex to handle obstacle avoidance, navigation, and so on~\cite{sanes2010design}. Neurons in the optic lobe react to brightness changes in the scene~\cite{posch2014retinomorphic}, thus resulting in a sparse and asynchronous coding and processing of the information.

The recent development of event-based cameras~\cite{gallego2019event}, for which pixels output events (i.e. spikes) whenever the brightness changes, and of spiking neural network (SNN) processors opens a perspective to mimic the extremely fast, energy-efficient, and robust nature of insects' visual navigation skills. Unfortunately, the use of SNNs in robotics remains marginal, limited by the synchronous properties of conventional processors. Furthermore, training SNNs is more difficult than training regular neural networks, therefore contributing to the neglect of spiking solutions for the benefit of conventional strategies. 

Over the past few years, extensive work has been done to design and build neuromorphic processors like HICANN~\cite{schemmel2010wafer}, NeuroGrid~\cite{benjamin2014neurogrid}, TrueNorth~\cite{merolla2014million} and Spinnaker~\cite{furber2014spinnaker}, or more recently SPOON~\cite{frenkel202028} (see~\cite{thakur2018large} for a review). The Loihi neuromorphic processor was first introduced in 2018 by Intel (\Cref{fig:main}A)~\cite{davies2018loihi}. The chip is fully digital and asynchronous, and features 128 neuromorphic cores, each containing 1,024 spiking neural units, for a total of 131,072 neurons and more than 130 million synapses. Programming Loihi is done by means of a Python-based API that allows to define the SNN topology and parameters, program learning rules, and setup communication with host system~\cite{lin2018programming}. Regarding its energy and time performance, Loihi demonstrates incredibly efficiency. The average energy per neuron update is ranging from 52~pJ (inactive) to 81~pJ (active), while the time per neuron update ranges from 5.3~ns to 8.4~ns respectively~\cite{davies2018loihi}. A comparison with the Intel Atom processor for solving an $\ell_1$ minimization problem clearly highlighted the advantage of Loihi over conventional CPUs, particularly for large problems. 

The emergence of neuromorphic processors like Intel's Loihi yields new promises for the applications of SNNs in robotics, particularly in super-constrained systems such as MAVs. In previous work, we evolved a spiking thrust controller in a highly abstracted simulation environment to control the landing of an MAV using the divergence of the ventral optic flow field~\cite{hagenaars2020evolved}. Autonomous landing of MAVs offers a simple yet challenging robotic use case for the development of neuromorphic controllers, first because of the difficulty to establish a fully operational, embedded hardware setup integrating the neuromorphic processor, but also because of the challenge introduced by the reality gap existing between state-of-the-art architectures proposed in the literature and benchmarked in simulation, and the constraints leveraged by neuromorphic chips. 

\begin{figure}[!t]
        \centering
        \includegraphics[width=0.6\linewidth]{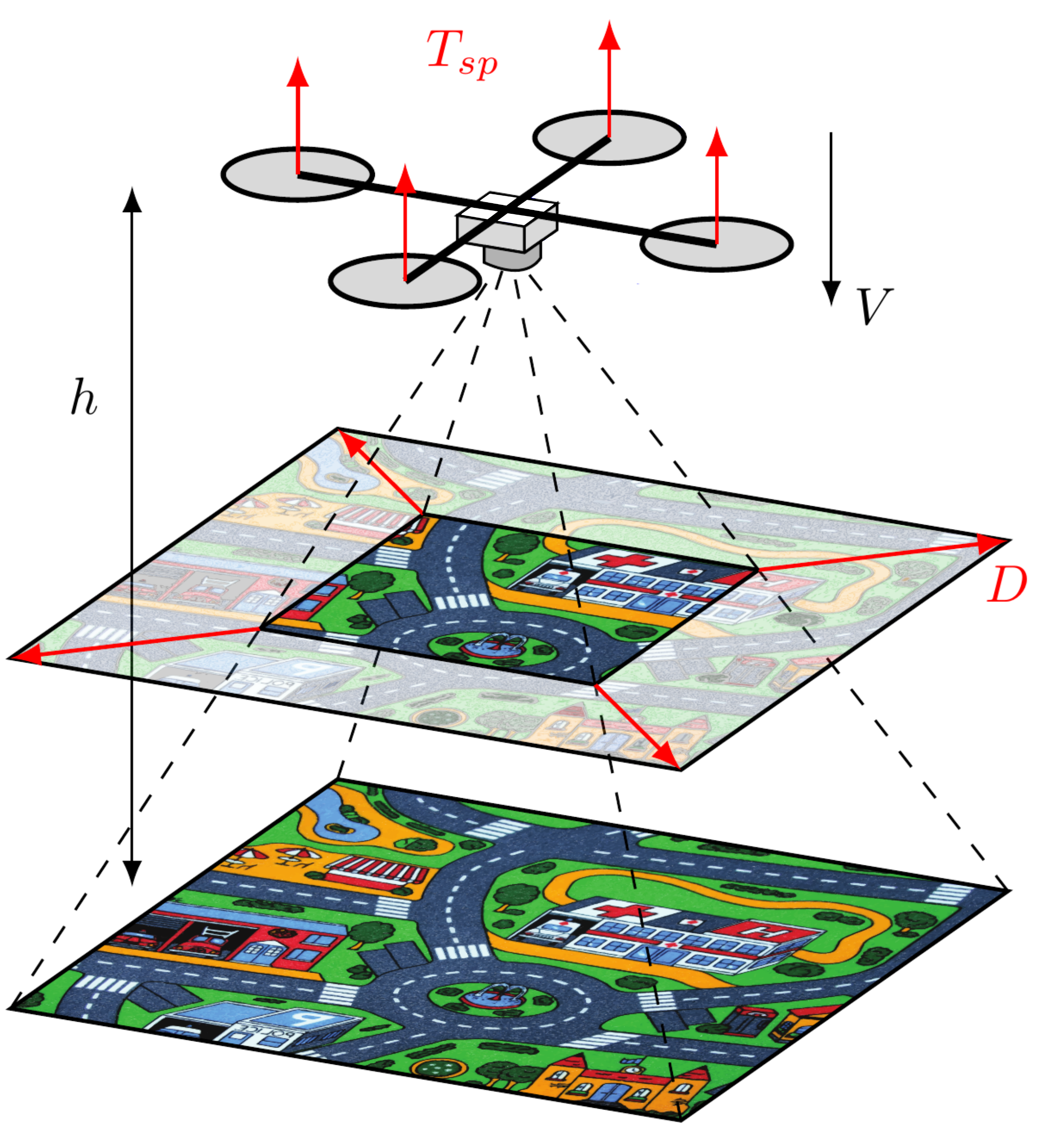}
        \caption{Overview of the proposed problem. A quadrotor equipped with a downward-facing camera is controlled through the embedded neuromorphic chip Loihi to perform autonomous vertical landings. The neuromorphic controller determines the thrust setpoint $T_{sp}$ based on the divergence $D$ of the optical flow field.}
        \label{fig:overview}
    \end{figure}

In this work, we introduce the very first fully embedded robotic application of the Loihi neuromorphic chip to control an MAV (\Cref{fig:main}). The SNN architecture is evolved in a highly randomized and abstracted vertical simulation and takes the ventral optic flow divergence as its input to determine the thrust setpoint to achieve a smooth landing (\Cref{fig:overview}). With this use case, we bridge the reality gap between simulated SNNs and neuromorphic hardware. Early robotic applications of Loihi are introduced in \Cref{sec:related_work}. In \Cref{sec:mat_meth}, we introduce the divergence estimator, the spiking neuron model, and the evolution of the SNN. Simulation and real-world results along with a quantitative analysis are then provided in \Cref{sec:results}.

\section{Related work}
\label{sec:related_work}

The recent introduction of neuromorphic processors is yielding brand new fields of research in robotics, ranging from low-level neuromorphic controllers to autonomous navigation, including vision-based obstacle avoidance. So far, a few attempts to integrate neuromorphic chips, including Loihi, already exist in the literature, yet most are restricted to simulation and offline testing. In~\cite{polykretis2020astrocyte}, a spiking model of the well-known central pattern generator was proposed to robustly control the walking of a hexapod robot. The network was tested on the Loihi chip in a ROS-based simulation. Besides, a spiking implementation of the anisotropic network has been recently achieved on the Loihi processor~\cite{michaelis2020robust}, showing that the chip reliably encodes neural activity patterns representing actions in the context of the motor control of a robotic arm. Using a sparse coding approach, the SNN allowed to control the angular position of a constrained 1-DoF unmanned aerial vehicle (UAV). In~\cite{kreiser2019error,kreiser2020error}, an SNN inspired by the head direction network of the fruit fly was proposed to implement an orientation estimator (i.e. compass) based on motor commands. Endowed with online learning, the model was simulated on the Loihi chip, thus paving the way towards a neuromorphic implementation of the so-called path integrator of insects. 

It is evident that both the reality gap between SNN simulation and neuromorphic hardware, as well as the hardware integration in a reliable real-time framework, form the two most important barriers towards robotic applications, even more so in the context of MAVs. However, a few encouraging works lately demonstrate the promising opportunities of neuromorphic hardware on board robots. For instance, a neuromorphic head pose estimation was performed with the Loihi chip and further tested in real-time robotic experiments on the iCub humanoid robot~\cite{renner2019event}. Efforts were also made in the field of autonomous mobile ground robots. In particular, autonomous outdoor navigation was achieved on IBM's TrueNorth hardware using deep convolutional neural networks on board a ground robot~\cite{hwu2017adaptive}. Also, neuromorphic navigation was achieved with the Turtlebot 2 with the Loihi chip~\cite{tang2020reinforcement}. After training a spiking deep deterministic policy gradient on Loihi in a Gazebo simulation, the robot successfully navigated in the real world, while showing a 75-fold increase in energy efficiency when compared to Jetson TX2. The use of neuromorphic hardware in the control of UAVs remains a difficult challenge due to the hardware constraints themselves (suitable embedded CPU, maximum payload, autonomy, etc.), but also due to the added complexity of controlling a 6-DoF UAV/MAV compared to a ground vehicle. The authors of~\cite{stagsted2020towards} recently unveiled the very first neuromorphic PID controller on hardware for UAVs, which was tested to control roll of a 1-DoF UAV. However, to our knowledge, there is no study reporting any fully embedded application of neuromorphic processors on board UAVs or MAVs.

\section{Materials and methods}
\label{sec:mat_meth}

    \subsection{Vertical landing with optic flow divergence}
    
    This work uses the optic flow formulation from~\cite{de2013optic}, which assumes a camera with its optical axis orthogonal to a static planar scene, as is the case in \Cref{fig:overview}. Moving the camera along this axis causes an optic flow to be perceived, whose divergence corresponds to the ratio of the axial velocity $V$ to the distance $h$ from camera to surface: $D = V / h$. A computationally efficient and reliable estimate of divergence $\hat{D}$ can be obtained from the relative, temporal variation in distance between pairs of tracked corners~\cite{ho2018adaptive}:
    
    \begin{equation}
        \hat{D}(t) = \frac{1}{N_D} \sum_{i=1}^{N_D} \frac{1}{\Delta t} \frac{l_i(t) - l_i(t - \Delta t)}{l_i(t - \Delta t)}
        \label{eq:div}
    \end{equation}

    \noindent where $N_D$ is the number of pairs considered, $\Delta t$ is the time step, and $l_i(t)$ is the distance between a pair of tracked corners at time $t$. Corners are found using the FAST corner detector~\cite{rosten2006machine} and tracked with a pyramidal Lucas-Kanade tracker~\cite{bouguet2001pyramidal}.
    
    A smooth landing can be achieved by keeping divergence constant during approach. Controllers in this work therefore receive as input the error with a divergence setpoint $\hat{D}(t) - D_{sp}$, where $D_{sp} = 1$~s$^{-1}$, and control the appropriate thrust through a setpoint $T_{sp}$. The abstraction offered by this approach helps with bridging the reality gap~\cite{scheper2020evolution}.
    
    \subsection{Spiking neural networks}
    
        \subsubsection{Simulation}
        \label{sssec:sim}
        
        The simulated spiking neural network (SNN) consists of leaky integrate-and-fire (LIF) neurons~\cite{stein1965theoretical}, meaning the membrane potential $v_i(t)$ can be described as follows:
        \begin{align}
            v_i(t) &= \tau_{v_i} \cdot v_i(t - \Delta t) + u_i(t) \label{eq:LIF-1} \\
            u_i(t) &= \sum_{j} w_{ij} s_j(t) \label{eq:LIF-2}
        \end{align}
        
        Each time step, $v_i(t)$ decays by a factor $\tau_{v_i} \in [0, 1]$. A synaptic input current $u_i(t)$ results from incoming spikes $s_j(t)$ multiplied by their corresponding synaptic weights. If the membrane potential is larger than a threshold $\theta_i$, a spike $s_i(t)$ is emitted and $v_i(t)$ is reset to $v_\mathit{rest} = 0$.
        
\begin{figure}[!t]
            \centering
            \includegraphics[width=\linewidth]{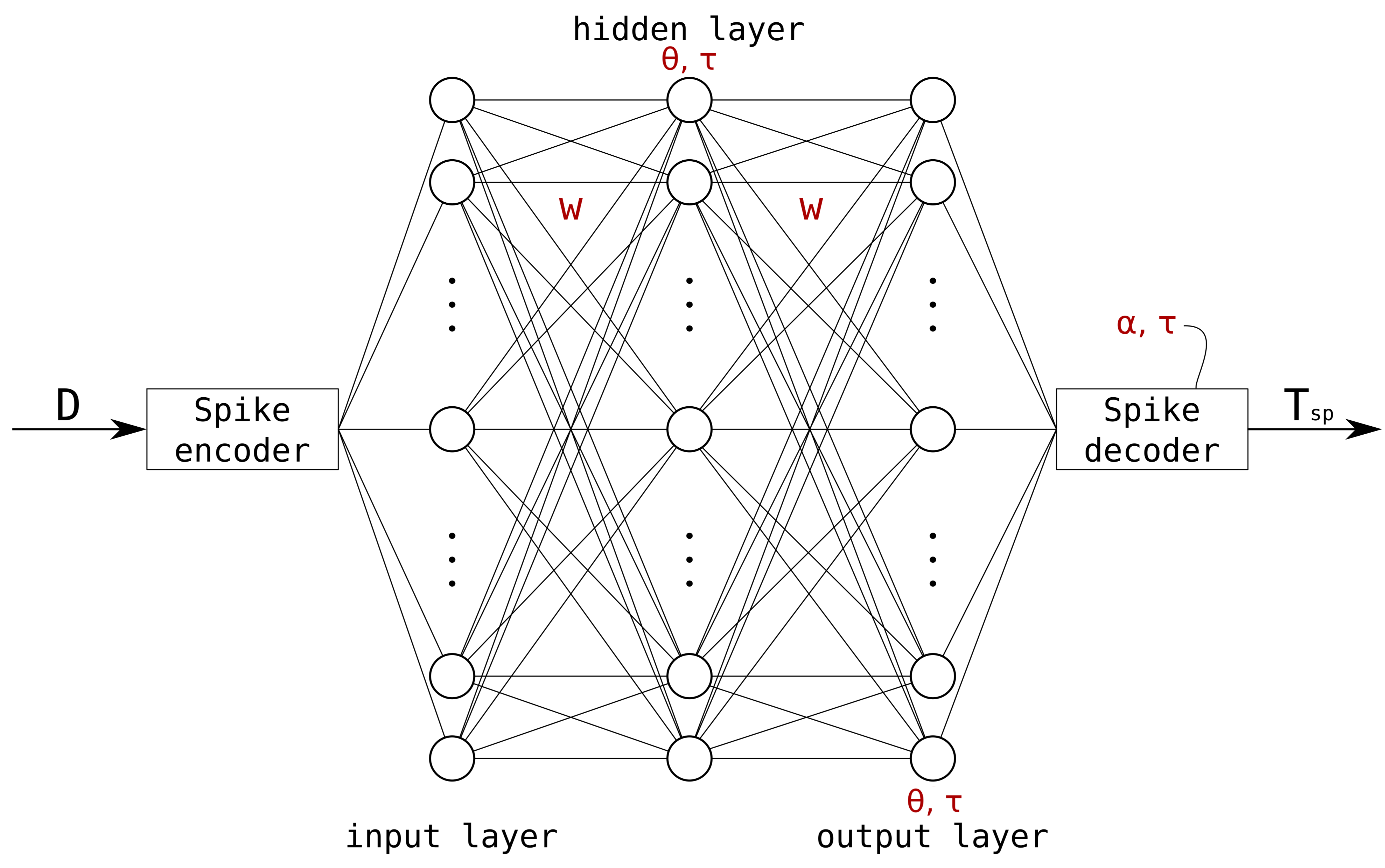}
            \caption{SNN controller architecture. Red variables $w$, $\alpha$, $\tau$, and $\theta$ are the evolved parameters. The hidden and output layer consist of 10 and 5 spiking LIF neurons respectively, following the dynamics described in \Cref{eq:LIF-1,eq:LIF-2}. The input layer of size 20 serves as a placeholder for spike transmission to the network.}
            \label{fig:snn}
        \end{figure}

        Our SNN controllers are made up of three layers, as can be seen in \Cref{fig:snn}. The hidden and output layers contain 10 and 5 LIF neurons respectively, while the input layer is just a placeholder of size 20 for input spikes produced by the spike encoder. We use a so-called position coding to convert a continuous signal to binary spikes: each value of $\hat{D}(t) - D_{sp}$ falls into one of 20 `buckets' and triggers a single spike. Buckets, or tuning curves, are distributed cubically with right-inclusive bounds $\frac{1}{10^2}\cdot\{-\infty, -10 \isep 10, \infty\}^3$~s$^{-1}$, where $-10 \isep 10$ indicates an integer interval.
        
        On the other side of the network, a decoding from spikes to a continuous $T_{sp}$ takes place. For this, we consider the spike trace $X_i(t)$ of each of the output neurons, which acts as a low-pass filter with decay $\tau_{x_i} \in [0, 1]$:
        \begin{equation}
            X_i(t) = \tau_{x_i} \cdot X_i(t - \Delta t) + \alpha_{x_i} \cdot s_i(t) 
            \label{eq:LIF-3}
        \end{equation}
        
        The thrust setpoint $T_{sp}$ is then taken as a weighted average of thrust vector $\mathbf{q}=\{q: q=-0.4+\frac{1}{5}n, n \in \{0\isep 4\}\}$~g (positive $T_{sp}$ upwards) and spike trace:
        \begin{equation}
            T_{sp}(t) = \frac{\sum_i^{} q_i \cdot X_i(t)}{\sum_i X_i(t)}
            \label{eq:thrust}
        \end{equation}
        
        \subsubsection{Loihi}
        
        The neurons implemented on Loihi follow a variation of the current-based (CUBA) LIF model~\cite{davies2018loihi}. Like our simulated LIF model, each neuron $i$ has two internal state variables: a synaptic input current $u_i(t)$ and a membrane potential $v_i(t)$:
        \begin{align}
            v_i(t) &= \tau_{v_i} \cdot v_i(t-1) + u_i(t) + b_i \label{eq:Loihi-1} \\
            u_i(t) &= \tau_{u_i} \cdot u_i(t-1) + 2^{6+\beta} \cdot \sum_j w_{ij} s_j(t) \label{eq:Loihi-2}
        \end{align}
        
        \noindent with current decay $\tau_{u_i} = (2^{12} - \delta_i^u) \cdot 2^{-12}$, voltage decay $\tau_{v_i} = (2^{12} - \delta_i^v) \cdot 2^{-12}$, voltage bias $b_i = 0$ and weight exponent $\beta = 0$. A spike $s_i(t)$ is emitted when the membrane potential reaches $\theta_i \cdot 2^6$, after which $v_i(t)$ is reset to $v_\mathit{rest} = 0$. The range and resolution of each parameter is given in \Cref{tab:prototypeParams}, all of them being integers.
        
        \renewcommand{\arraystretch}{1.2}
        \begin{table}[!h]
            \centering
            \caption{Range and resolution of neuron parameters on Loihi}
            \label{tab:prototypeParams}
            \begin{tabular}{l|cc}
                \hline\hline
                \textbf{Parameter} & \textbf{Range} & \textbf{Resolution} \\ \hline
                $w_{ij}$ & $[-256 \isep 254]$ & 2 \\
                $\theta_i$ & $[0 \isep 131071]$ & 1 \\
                $\delta_i^u$, $\delta_i^v$ & $[0 \isep 4096]$ & $1$ \\ \hline\hline
            \end{tabular}
        \end{table}
        
        \subsubsection{From simulated to on-chip spiking networks}
        
        Comparing \Cref{eq:LIF-1,eq:LIF-2} with \Cref{eq:Loihi-1,eq:Loihi-2} reveals that the on-chip neurons can be mimicked in simulation with several adaptations. Apart from having the parameters in \Cref{tab:prototypeParams} in the integer domain, we make sure that the synaptic current is reset at each time step by setting $\delta_i^u = 4096$, giving $\tau_{u_i} = 0$. Furthermore, we limit the range of the firing threshold $\theta_i$ to $[1 \isep 1024]$, both to prevent the simulated neurons from spiking when no input spikes are received (which does not happen on Loihi), and to allow spiking with fewer inputs.
        
        Additionally, we have to account for the way in which the network is updated. On Loihi, each group of neurons (i.e. layer) is only updated once per forward pass, meaning that an input spike would take $N$ time steps to reach the output layer, with $N$ the number of layers. This is not necessarily the case for the simulated SNN, but the phenomenon can be replicated by implementing delay buffers between layers.

        \subsubsection{Evolutionary optimization}
        
        We use a variant of the evolutionary strategy proposed in~\cite{scheper2020evolution,hagenaars2020evolved}, where parameters are evolved through mutation only, and evaluation occurs in a highly randomized and abstracted vertical simulation environment that is resampled every generation. The fitness of an individual is quantified as the accumulated divergence error:
        \begin{equation}
            \sum_t\,\,\abs{\hat{D}(t) - D_{sp}}
            \label{eq:fit}
        \end{equation}
        
        \noindent averaged over landings from four different starting heights $h_0 = \{2, 3, 4, 5\}$~m, with the individual controlling the thrust setpoint $T_{sp}$ of a point mass (resembling a drone) based on an observation of the divergence error. The simulation environment contains various and varying sources of noise and delay, as well as changing thrust dynamics, which helps with bridging the reality gap as it stimulates controllers that act independent of the exact settings (see~\cite{hagenaars2020evolved}).
        
        \renewcommand{\arraystretch}{1.2}
        \begin{table}[!h]
            \centering
            \caption{Parameters mutated during evolution}
            \begin{tabular}{l|cc}
                \hline\hline
                \textbf{Parameter} & \textbf{Range} & \textbf{Mutation} \\ \hline 
                $w_{ij}$ & $[-256 \isep 254]$ & $\mathcal{U}\{-84,84\}$ \\
                $\theta_i$ & $[1 \isep 1024]$ & $\mathcal{U}\{-341,341\}$ \\
                $\delta_i^u$, $\delta_i^v$ & $[0 \isep 4096]$ & $\mathcal{U}\{-1365,1365\}$ \\
                $\alpha_{x_i}$, $\tau_{x_i}$ & $[0,1]$ & $\mathcal{U}(-1/3, 1/3)$ \\ \hline\hline
            \end{tabular}
            \label{tab:mutatedParams}
        \end{table}
        
        Evolution starts off with a population of 100 randomly initialized individuals that undergo an initial evaluation. Each generation, an offspring of equal size (100) is created by selecting and duplicating the 50 fittest individuals, and mutating their parameters with $P_\mathit{mut} = 0.3$ according to \Cref{tab:mutatedParams}. In order to stimulate robustness to varying environmental conditions, both the parent population and its offspring are re-evaluated, after which the 100 fittest individuals are selected to become the next generation, of which we select and duplicate the 50 fittest to create offspring, etc.
        
    \subsection{Hardware integration of Loihi}

\begin{figure*}[!t]
            \centering
            \includegraphics[width=\textwidth]{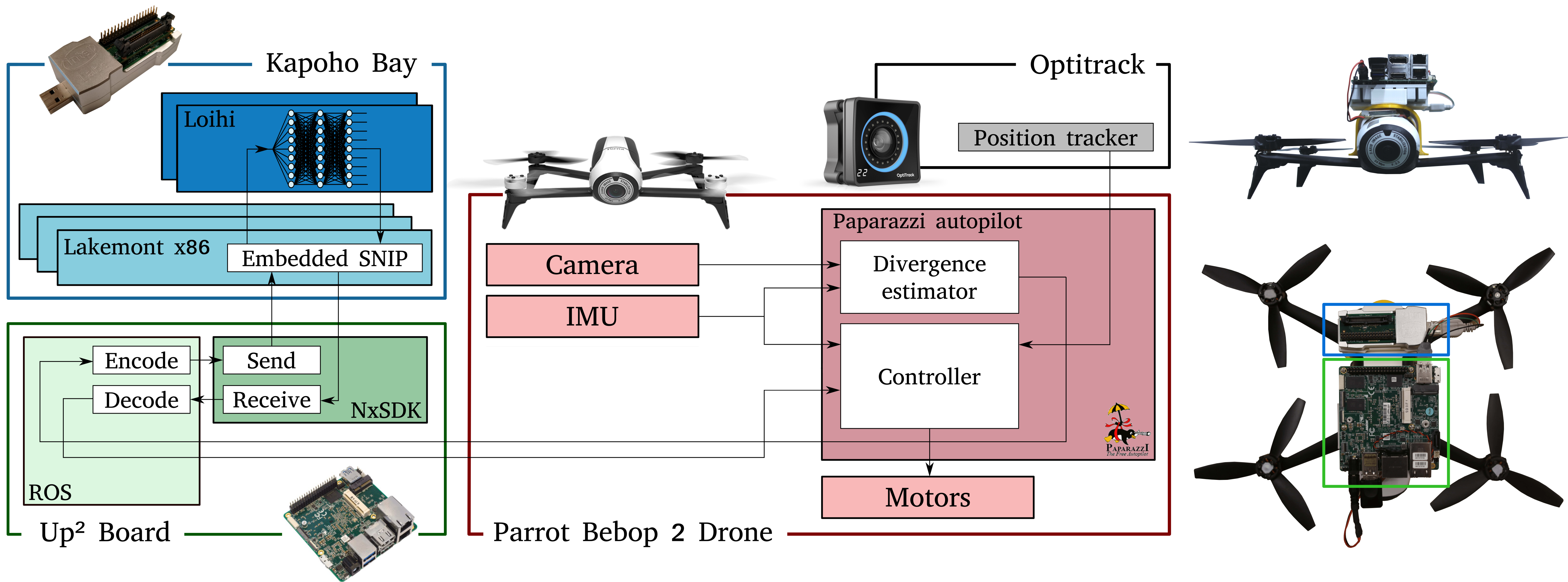}
            \caption{Hardware overview of the project. \textbf{Left:} The quadrotor, a Parrot Bebop 2, is flashed with a custom version of the Paparazzi autopilot software. A UART communication protocol is implemented to exchange divergence error and thrust setpoint between the UP$^2$ board and the drone. Spike encoding and decoding is performed within the ROS environment, interfaced with the Intel NxSDK to ensure proper spike exchange with the Kapoho Bay. The Optitrack motion capture system provides the drone with position data to control its $(x,y)$ position. \textbf{Right:} Front and top view of the Parrot Bebop 2 drone equipped with the Kapoho Bay. \textit{Blue:} Intel Kapoho Bay neuromorphic device. \textit{Green:} UP$^2$ board.}
            \label{fig:architecture}
        \end{figure*}

        In this study, we use Intel's Kapoho Bay neuromorphic device (\Cref{fig:architecture}), which has a USB-stick form factor and which contains two Loihi chips for a total of 256 neuro-cores, able to represent 262,144 neurons and 260 million synapses. Besides, each chip incorporates three embedded CPUs (Lakemont x86) used for monitoring and I/O spike management. 
        
        Working with the current version of the Kapoho Bay requires a specific hardware setup. For this reason, it was decided to embed the neuromorphic device on a UP$^2$ (UP Squared) board, which features an Intel Atom x7-E3950 processor (up to 2.00 GHz) along with 4 GB of RAM. The OS is a standard Linux Ubuntu 16.04, with ROS Kinetic, Python 3.5.2 and version 0.9.5 of the Intel NxSDK. The left part of \Cref{fig:architecture} gives an overview of this. 
        
        Spike encoding and decoding of respectively the divergence error $\hat{D}(t) - D_{sp}$ and the thrust setpoint $T_{sp}$ happens within the ROS environment, following the methods described in \Cref{sssec:sim}. Generated spikes are transmitted to the embedded CPU (Lakemont x86) of the Kapoho Bay through a pipeline established with the NxSDK. There, an embedded SNIP (sequential neural interacting process) manages the communication of the spikes to and from the neuro-cores on Loihi, and executes the SNN for the current time step. The output spikes are then sent to the UP$^2$ board using the same channel, then further decoded into a thrust setpoint sent to the Paparazzi autopilot over UART. 
        
\section{Results}
\label{sec:results}

    \subsection{Experimental setup}
    
        \subsubsection{Simulation}
        
        We performed four randomly initialized evolutions of 200 generations and analyzed the sensitivity of the final generations in terms of time to land and touchdown velocity. A robust individual was then selected for further testing.
        SNNs were simulated using Python and the open-source library PySNN\footnote{\url{https://github.com/BasBuller/PySNN}} recently developed in our lab. Evolutionary optimization was performed with the DEAP~\cite{fortin2012deap}. 
        
        \subsubsection{Real world}
        
        Real-world flight tests were performed with a Parrot Bebop 2 quadrotor featuring a 780 MHz dual-core ARM Cortex A9 processor running the open-source Paparazzi\footnote{\url{https://github.com/paparazzi/paparazzi}} autopilot. Landings start from an initial altitude of 4.0~m and are ended at 0.1~m. During the landing of the drone, only the vertical motion is controlled with the Loihi chip. The $(x,y)$ position of the drone is controlled using the OptiTrack motion capture system. The divergence of the optic flow field is estimated on-board the drone based on the output of the downward-facing CMOS camera and derotated using the IMU, at a rate of approximately 60~Hz. Thrust setpoints received by the autopilot over UART are converted to motor commands using a PI controller (with gains $P = 5.0$, $I = 0.3$), in order to account for unmodeled aerodynamics~\cite{scheper2020evolution}. See \Cref{fig:architecture} for a hardware overview.  
        
    \subsection{Simulation results}
    
\begin{figure}[!h]
            \centering
            \includegraphics[width=0.9\linewidth]{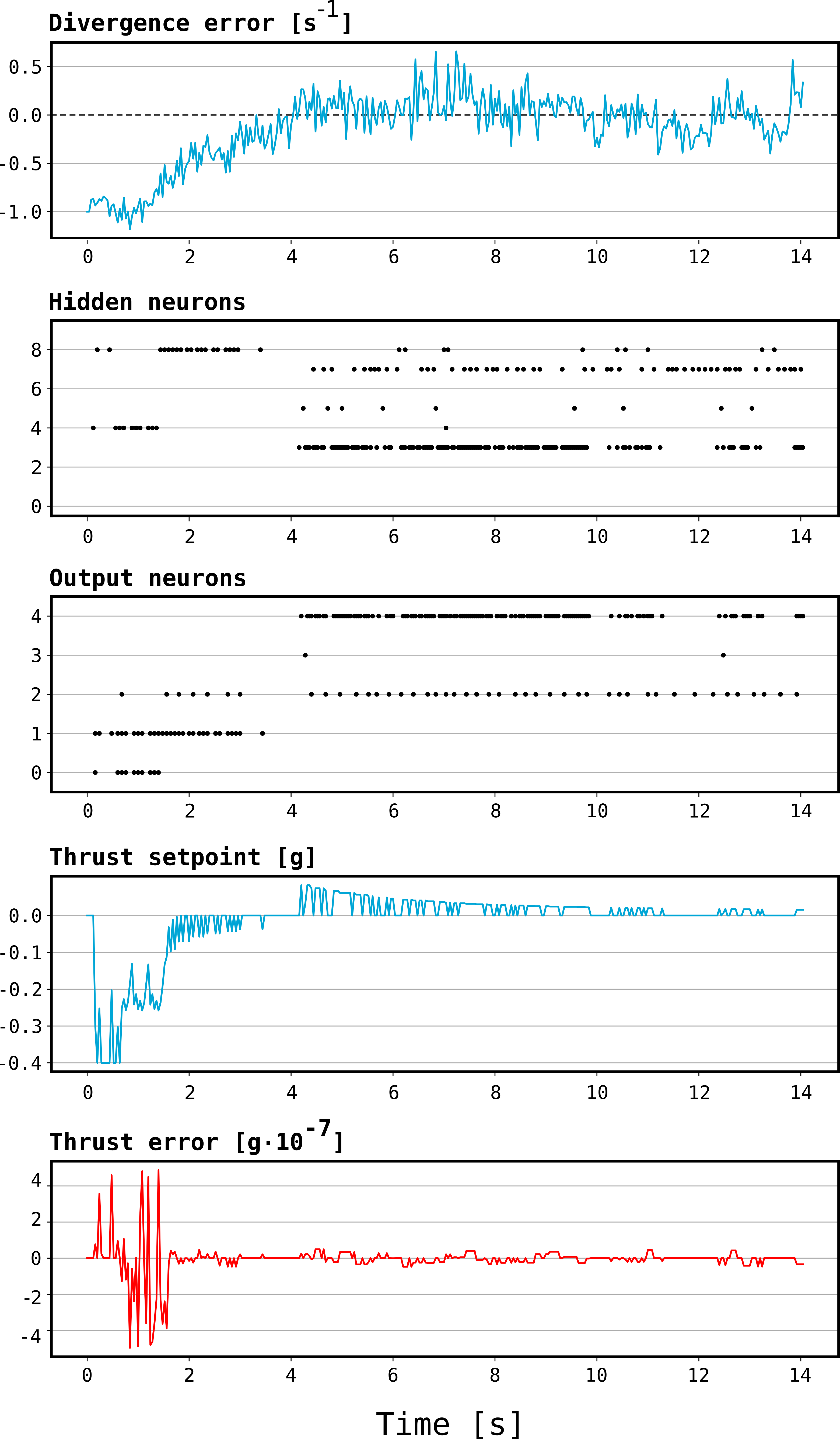}
            \caption{One of the results in simulation obtained with the Loihi chip. The divergence error is fed to both the simulated SNN (PySNN) and the on-chip SNN (Loihi). Hidden and output layer spikes sequences are given, as well as the thrust setpoint decoded from the output spikes. The error between the simulated SNN and the on-chip SNN is shown on the bottom chart.}
            \label{fig:sim_res_sample}
        \end{figure}

        The selected individual was put through a set of 100 simulated landing runs from $h_0 = 4$~m in a randomized environment to assess performance. We recorded the input to the SNN (divergence error) and its output (thrust setpoint), as well as the spikes in both the hidden and output layer. Next, we replaced the network implemented in PySNN with one running on Loihi, allowing us to examine the reality gap between the two by feeding the same inputs and comparing spikes. \Cref{fig:sim_res_sample} shows this comparison for a single landing.

        For each landing run, we compare spike sequences with two metrics: the infill percentage and the matching score. In the following, we define the infill percentage of a spiking sequence as the ratio of the number of spikes within a layer to the maximum of spikes of that layer: 

        \begin{equation}
            \mathrm{Infill} = \frac{\sum_{t=0}^{T-1} \sum_{i=0}^{N_n-1} s_i(t)}{T \cdot N_n}
            \label{eq:infill}
        \end{equation}
        
        \noindent where $T$ is the total number of time steps, and $N_n$ is the number of neurons in the layer. Similarly, we define the matching score between PySNN and Loihi as follows:
        
        \begin{equation}
            \mathrm{Match} = 1 - \frac{\sum_{t=0}^{T-1} \sum_{i=0}^{N_n-1} \abs{s_i^{\mathit{PySNN}}(t) - s_i^{\mathit{Loihi}}(t)}}{T \cdot N_n}
            \label{eq:match}
        \end{equation}
    
        The resulting scores are given in \Cref{tab:sim_res}. When comparing between PySNN and Loihi, we note excellent matching scores, $99.8\% \pm 0.3\%$ (mean $\pm$ standard deviation, s.d.) in the hidden layer, and $99.7\% \pm 0.7\%$ in the output layer. These correspond do coefficients of variations $C_v = \sigma / \mu$ as low as 0.003 and 0.007 respectively. Results also show that the average infill is only $6.0\%$ in the hidden layer, and $10.6\%$ in the output layer. Statistical analysis of the infill reveals a normal distribution (Kolmogorov-Smirnov normality test, $p \gg 0.05$), while the two distributions (PySNN and Loihi) were found statistically not different (two-sided Kolmogorov-Smirnov test, $p = 1.00$) for all network layers. All together, these results demonstrate that there are no significant differences between the behaviors exhibited by the simulated (PySNN) and the on-chip (Loihi) spiking networks. 
        
            \begin{table}[!h]
                \centering
                \caption{Spike infill and matching for PySNN and Loihi}
                \begin{tabular}{l|rrr|rrr}
                    \hline\hline
                     & \multicolumn{3}{c|}{\textbf{Hidden layer ($N = 100$)}} & \multicolumn{3}{c}{\textbf{Output layer ($N = 100$)}} \\ \hline
                     & \multicolumn{1}{c}{PySNN} & \multicolumn{1}{c}{Loihi} & \multicolumn{1}{l|}{\textit{Match}} & \multicolumn{1}{c}{PySNN} & \multicolumn{1}{c}{Loihi} & \multicolumn{1}{l}{\textit{Match}} \\ \hline
                    \textit{mean} & $6.0\%$ & \multicolumn{1}{r|}{$6.0\%$} & $99.8\%$ & $10.7\%$ & \multicolumn{1}{r|}{$10.6\%$} & $99.7\%$ \\
                    \textit{med.} & $6.0\%$ & \multicolumn{1}{r|}{$6.0\%$} & $99.9\%$ & $10.6\%$ & \multicolumn{1}{r|}{$10.5\%$} & $99.9\%$ \\
                    \textit{s.d.} & $0.5\%$ & \multicolumn{1}{r|}{$0.5\%$} & $0.3\%$ & $1.1\%$ & \multicolumn{1}{r|}{$1.1\%$} & $0.7\%$ \\
                    \textit{min} & $5.0\%$ & \multicolumn{1}{r|}{$5.0\%$} & $97.9\%$ & $8.3\%$ & \multicolumn{1}{r|}{$8.3\%$} & $95.7\%$ \\
                    \textit{max} & $7.5\%$ & \multicolumn{1}{r|}{$7.5\%$} & $100.0\%$ & $13.8\%$ & \multicolumn{1}{r|}{$13.8\%$} & $100.0\%$ \\ \hline\hline
                \end{tabular}
                \label{tab:sim_res}
            \end{table}
            
        A quantitative comparison of the resulting thrust setpoint sequences was also conducted by means of the root-mean-square error (RMSE) between PySNN and Loihi. The average RMSE was found equal to $0.0050 \pm 0.0069$~g, thus confirming the equivalence between the PySNN simulation and the Loihi implementation. The error between the two output thrust setpoint sequences is provided in \Cref{fig:sim_res_sample}.
        
        According to these results, it is clear that the reality gap both simulation (PySNN) and hardware (Loihi) has been bridged through our evolutionary approach, thus allowing further real robotic applications as introduced below. 
        
    \subsection{Real-world landings with Loihi}
    
        Next, we performed 20 real-world landings with the SNN controller implemented on Loihi. Again, network input (divergence error), output (thrust setpoint) and spikes were recorded. The OptiTrack system available in our flying arena was used to track vertical speed and position during landing. 
        
\begin{figure}[!t]
            \centering
            \includegraphics[width=\linewidth]{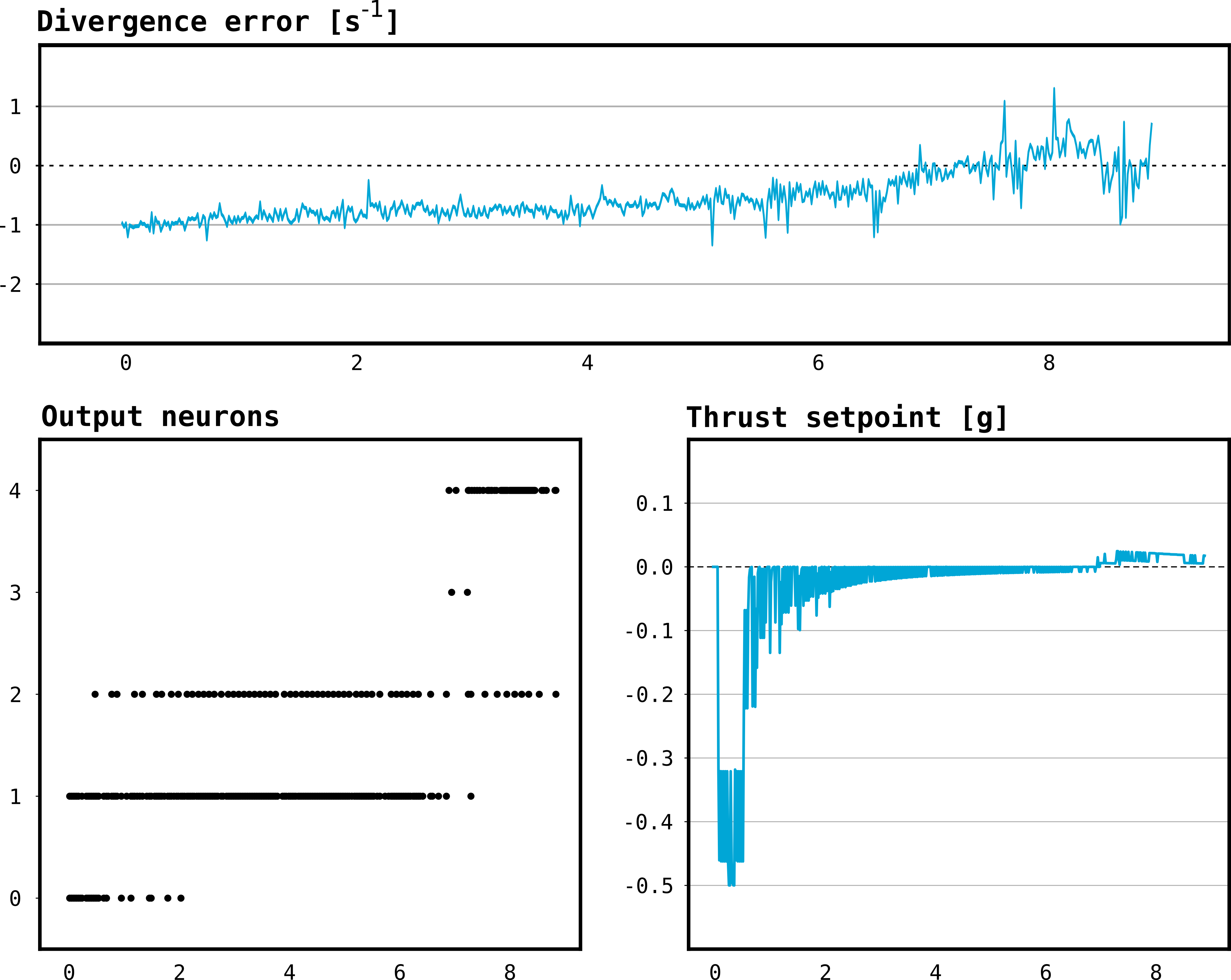}
            \caption{Typical network input and output for a real-world landing performed with the Loihi neuromorphic chip. The divergence error is fed to the SNN, eventually resulting in spikes in the five output neurons, and in a thrust setpoint that is sent to the drone.}
            \label{fig:landing_loihi}
        \end{figure}

\begin{figure}[!t]
            \centering
            \includegraphics[width=\linewidth]{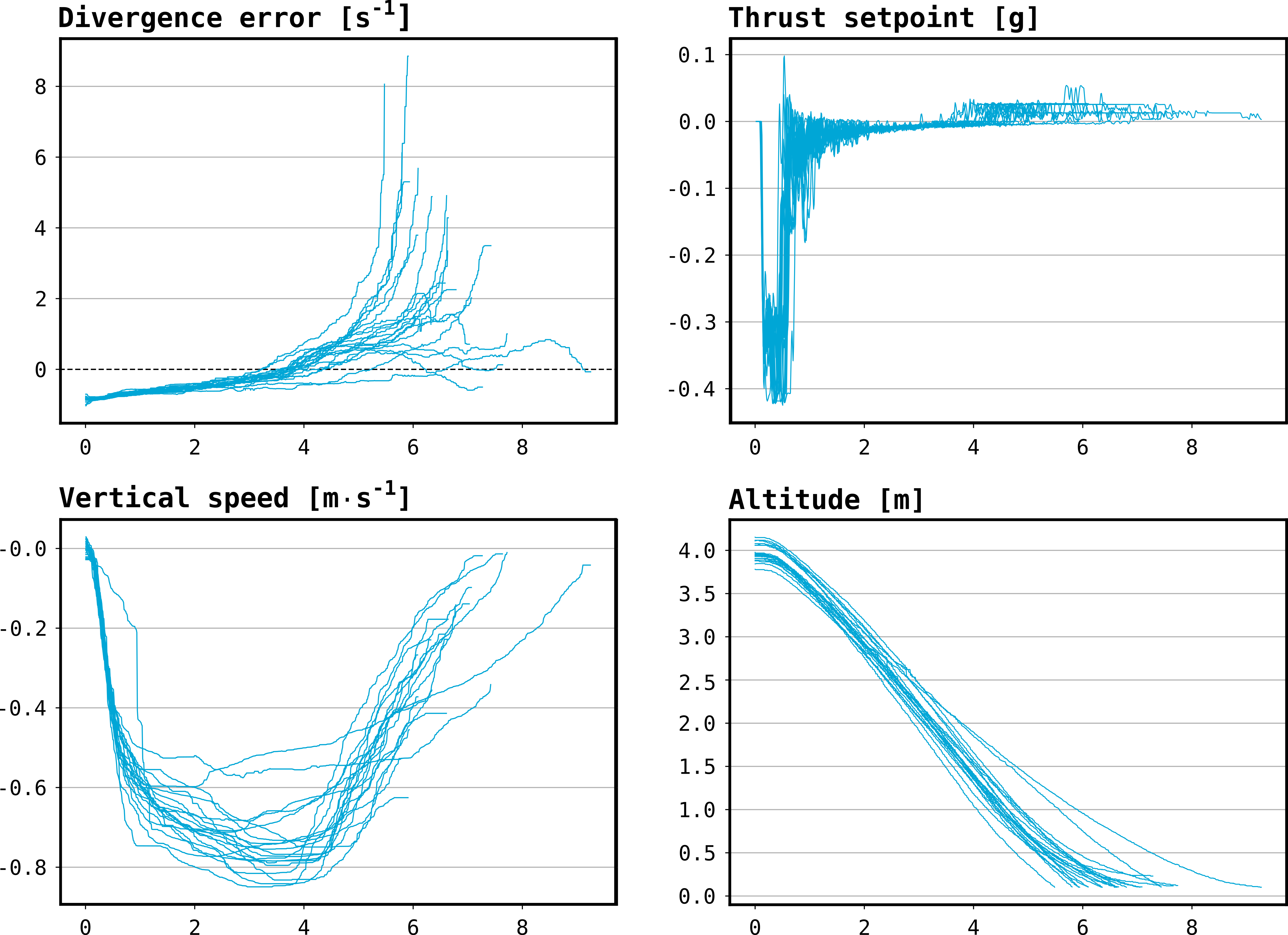}
            \caption{Overall results of the landing tests with the Loihi neuromorphic chip ($N=20$). All tests started at height of 4.0~m.}
            \label{fig:landing_results}
        \end{figure}

        \Cref{fig:landing_loihi} shows the output spikes and decoded thrust setpoint for a single real-world landing. The infill percentage of the output layer was determined for all runs, giving an average infill of $13.6\% \pm 0.9\%$, also following a normal distribution (Kolmogorov-Smirnov normality test, $p \gg 0.05$). Nevertheless, there still remain some differences between landings performed with Loihi in simulation and the real world. For instance, it seems that a braking thrust (positive setpoint) to prevent a positive divergence error is applied later in the real world (\Cref{fig:landing_loihi}, $t = 7$~s) than in simulation (\Cref{fig:sim_res_sample}, $t = 4$~s). This could be caused by an additional hardware delay. All landing runs were performed in a very smooth fashion, as observed in \Cref{fig:landing_results}. The SNN controller implemented on Loihi performs consistently and roughly takes 6-8 seconds to safely land the drone. Videos of the flight tests can be found at \url{https://mavlab.tudelft.nl/loihi/}. Lastly, we monitored the execution time of the SNN per timestep. Results show that the average execution time is as low as $3.78 \pm 0.07$~\textmu s. 
        
\section{Conclusions}

    We implemented an SNN for visual-based thrust control of a quadrotor on Intel's Loihi neuromorphic research processor. Using an evolutionary algorithm to optimize SNN controllers and an abstracted simulation environment to evaluate them, we closed the reality gap between simulated and on-chip SNNs, with a resulting controller consisting of only 35 neurons. During real-world landing tests from an altitude of 4~m, the controller proved robust and consistent. To the best knowledge of the authors, this work presents the very first embedded application of Loihi in a flying robot. Future work will involve the use of the developed software to further investigate the performance of Loihi with MAVs.  

\section*{Acknowledgments}

    The authors would like to thank Nilay Sheth and Alpha Renner for their support with the hardware setup. The authors are also grateful to the Intel Neuromorphic Computing Lab and the Intel Neuromorphic Research Community for their support with the Kapoho Bay.

\bibliographystyle{IEEEtran}
\bibliography{biblio}

\end{document}